# Dimensionality Reduction in Deep Learning for Chest X-Ray Analysis of Lung Cancer

Yu. Gordienko[*], Yu. Kochura, O. Alienin,
O. Rokovyi[1], and S. Stirenko[1]
National Technical University of Ukraine,
"Igor Sikorsky Kyiv Polytechnic Institute",
Kyiv, Ukraine
yuri.gordienko@gmail.com

Peng Gang, Jiang Hui, Wei Zeng
Huizhou University,
Huizhou City, China
peng@hzu.edu.cn

*Abstract*—Efficiency of some dimensionality reduction techniques, like lung segmentation, bone shadow exclusion, and t-distributed stochastic neighbor embedding (t-SNE) for exclusion of outliers, is estimated for analysis of chest X-ray (CXR) 2D images by deep learning approach to help radiologists identify marks of lung cancer in CXR. Training and validation of the simple convolutional neural network (CNN) was performed on the open JSRT dataset (dataset #01), the JSRT after bone shadow exclusion — BSE-JSRT (dataset #02), JSRT after lung segmentation (dataset #03), BSE-JSRT after lung segmentation (dataset #04), and segmented BSE-JSRT after exclusion of outliers by t-SNE method (dataset #05). The results demonstrate that the pre-processed dataset obtained after lung segmentation, bone shadow exclusion, and filtering out the outliers by t-SNE (dataset #05) demonstrates the highest training rate and best accuracy in comparison to the other pre-processed datasets.

## I. INTRODUCTION

Nowadays, chest X-ray (CXR) imaging is used widely for health monitoring and diagnosis of many lung diseases (pneumonia, tuberculosis, cancer, etc.), because of relatively low price and affordability. Usually, detection by CXR of marks of these diseases, especially cancer, is performed by expert radiologists, which is long and complicated process. It is delayed by cumbersome manual analysis and limited by a shortage of specialists. For example, in China, the annual number of diagnosed lung cancer cases is very big (~half million), but the quantity of the certified radiologists is much lower (~$10^5$) for nationwide screening of all population (~$10^9$) [1]. But the current evolution of numerical computing for analysis of medical images [2], including machine learning and deep learning [3] techniques, allows scientists by means of, for example, CheXNet model to detect automatically many lung diseases from CXR images at a level exceeding certified radiologists [4]. That is why the new computing tools and machine learning techniques will be very important for the faster and better medical image analysis for the subsequent diagnostic. The main aim of this paper is to demonstrate efficiency of dimensionality reduction performed by lung segmentation, bone shadow exclusion, and t-distributed stochastic neighbor embedding (t-SNE) techniques for analysis of 2D CXR of lung cancer patients.

## II. PROBLEM AND RELATED WORK

Among various screening methods used to detect suspicious lung cancer marks, CXR is often thought as an obsolete medical imaging method. But recently, promising results were obtained in the field of lung diseases diagnostic by deep learning in relation to CXR image analysis [4,5]. Several open datasets with CXR images allowed data scientists to train, verify, and tune their new deep learning algorithms. For example, Japanese Society of Radiological Technology (JSRT) opened access to their database of CXR images with and without lung cancer nodules (Fig. 1a) [6].

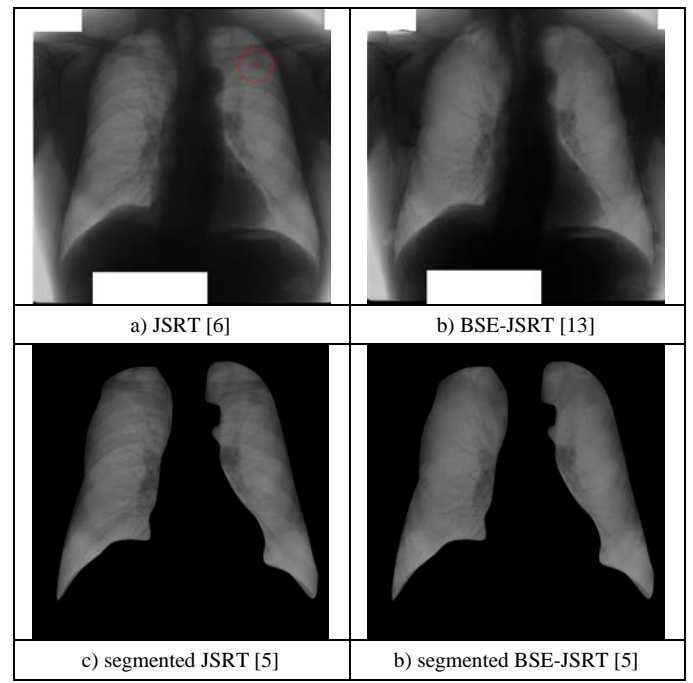

| a) JSRT [6] | b) BSE-JSRT [13] |
| c) segmented JSRT [5] | b) segmented BSE-JSRT [5] |

Fig. 1. Example of the original image (2048×2048 pixels) with the cancer nodule from JSRT dataset [6] (a), the correspondent image without bones from BSE-JSRT dataset [13] (b), its segmented version of image from JSRT datanase [5] (c), and segmented version of the image from BSE-JSRT database [5] (d). The nodule location and region are denoted by the point and circle respectively (a).

Lung Image Database Consortium (LIDC) proposed the open database of images captured by various methods including dataset of ~$10^5$ CXR images [7]. The U.S. National Library of Medicine has made two open databases of CXR images to improve computer-aided diagnosis of pulmonary diseases [8,9]. Montgomery County (MC) database of CXR images was created in collaboration with the Department of Health and Human Services, Montgomery County in Maryland, USA. Shenzhen Hospital (SH) database of CXR images was acquired from Shenzhen No. 3 People's Hospital in Shenzhen, China. Both datasets contain normal and abnormal CXR images with marks of tuberculosis. To the moment ChestX-ray14 is the largest publicly available database of CXR images. It contains more than $10^5$ CXR images with marks of 14 different lung diseases [10]. The CheXNet model achieved the promising results, which were mentioned above, by training on ChestX-ray14 dataset.

Several problems with computer-aided diagnosis of lung diseases (especially with detection of cancer marks) on the basis of these databases are related with presence of high variability of lungs and other regions outside of lungs (heart, shoulders, ribs, clavicles, etc.). This variability can be explained by different gender, nationality, age, substance abuse, professional activity, general health state, and other parameters of patients under investigations. Also it can be enhanced by artifacts like notes or labels created by doctors on analogous photos (like white rectangles in places of such notes shown in top and bottom parts on Fig. 1a and Fig. 1b). Unfortunately, the representativeness and homogeneity levels of databases with regard to these parameters are not considered and reported enough. It is very sensitive question for reliability of any predictions on the basis of these datasets, especially for the small datasets like JSRT and BSE-JSRT.

To increase the accuracy of predictions, the outside regions which are not pertinent to lungs or other regions of interest were proposed to be excluded in some research. For this purpose "external segmentation" of the left and right lung fields in CXR images was investigated. Recently, various segmentation methods were considered which are based on active shape models, active appearance models, and a multi-resolution pixel classification method [11,12]. "Internal segmentation" is related with diminishing the effect of some body parts that shadow the lung, for example, ribs and clavicles. Chest Diagnostic System Research Group (Budapest, Hungary) provided the bone shadow eliminated (BSE) version of JSRT database (BSE-JSRT) [13]. BSE-JSRT database contains 247 images (Fig. 1b) of the JSRT dataset, where shadows of ribs and clavicles were thoroughly removed from CXR images of JSRT database by the special algorithms.

Despite these attempts, the body part segmentation remains a challenge in medical image applications, especially in the view of high variability of lungs and other regions outside of lungs due to different personal parameters of patients under investigations. The one of the aims of the work was to check the difference between application of the deep learning approach to the original JSRT dataset (below it is mentioned as dataset #01), BSE-JSRT dataset, i.e. the same JSRT dataset, but without clavicle and rib shadows (dataset #02), and their segmented versions, namely, segmented JSRT dataset (dataset #03) (Fig.1c), segmented BSE-JSRT dataset (dataset #04) (Fig.1d), and segmented BSE-JSRT after exclusion of outliers by t-SNE method (dataset #05)

## III. DATA AND METHODS

### A. Exploratory Data Analysis of Dataset

JSRT and BSE-JSRT database contains 247 images each: 154 cases with lung nodules and 93 cases without lung nodules [6]. The exploratory data analysis shown that this small dataset is not well-balanced with regard to availability/absence of nodule, age, type of nodule (malignant/benign), size of nodule (Fig.2), degree of subtlety (Fig.3), combined distribution among gender-size(of nodule)-degree(of subtlety) (Fig.4).

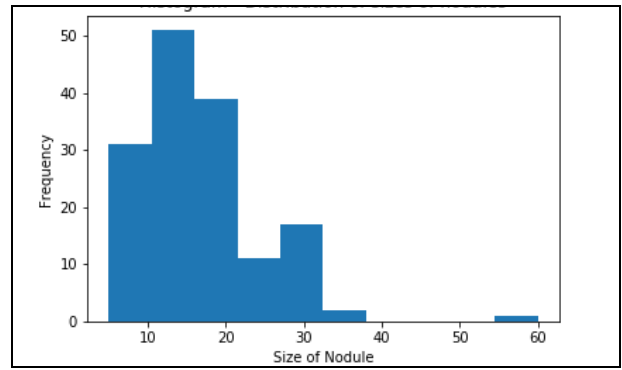

Fig. 2. Distribution of nodule sizes among JSRT images.

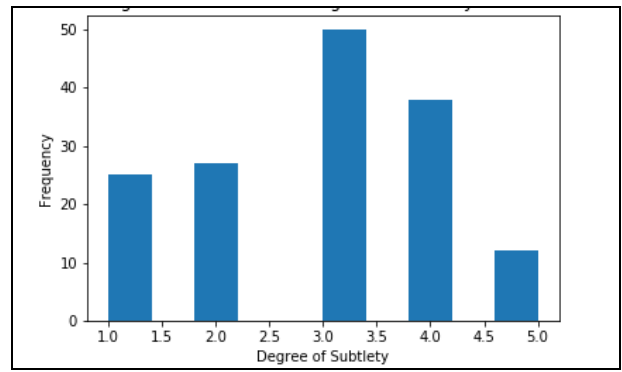

Fig. 3. Distribution of degrees of subtlety among JSRT images.

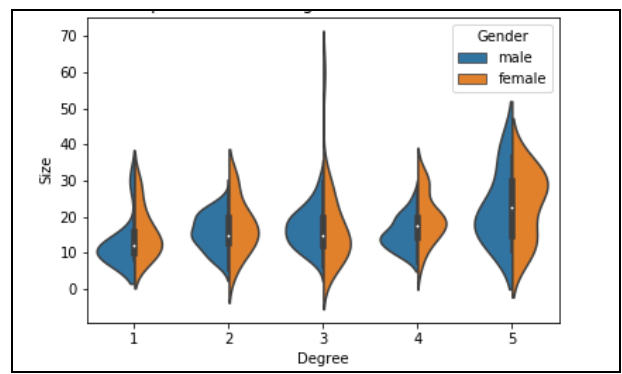

Fig. 4. Combined distribution of genders, sizes of nodules, and degrees of subtkety among JSRT images.

The locations of nodules inside lungs are shown in Fig.5.

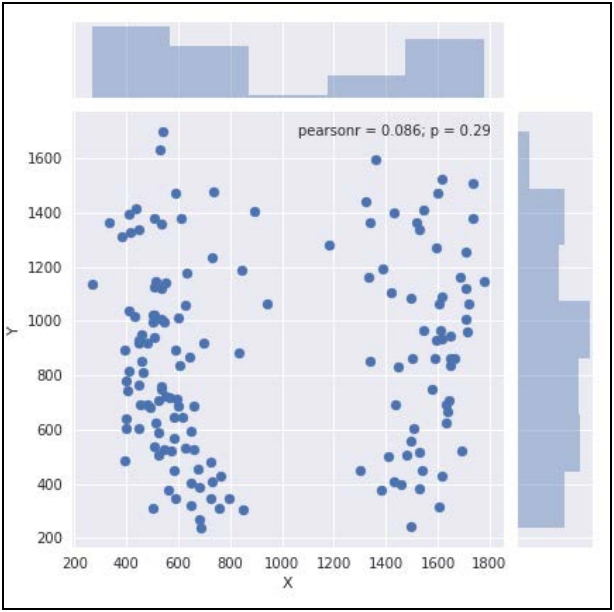

Fig. 5. Locations of nodules by X and Y coordinates (in pixels) with regard to image sizes.

### B. Dimensionality Reduction by Segmentation and Bone Shadows Exclusion

To perform lung segmentation, i.e. to cut the left and right lung fields from the lung parts in standard CXRs, the manually prepared masks were used (Fig.6) that allowed to get two additional datasets: segmented JSRT dataset (dataset #03) (Fig.1c), and segmented BSE-JSRT dataset (dataset #04) (Fig.1d). The high variability of lungs was observed for lung shape and borders and even without taking into account internal and external parts of lungs. Some examples of the most similar and dissimilar lung masks are shown in Fig.6. Unfortunately this variability cannot be excluded by any universal lung mask obtained by union, intersection, or mean operation on masks, because in all these cases some nodule locations (Fig.5) cannot be enclosed by the universal lung mask.

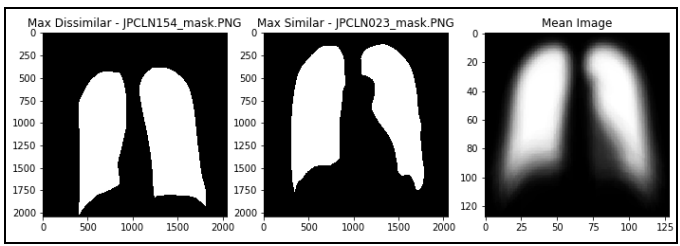

Fig. 6. Examples of the most dissimilar (left), similar (central), and mean (right) lung masks.

Nevertheless, the average portion of segmentation area was 0.32±0.06 (Fig.8) that allowed us to get dimensionality reduction by 3.2±0.7 times. The additional operation of bone shadow exclusion (Fig.7) was equivalent to exclusion of non-relevant features (bone shadows) that have the average portion of 0.15±0.04 (Fig.8) and give additional dimensionality reduction by ~2 times. Despite, the last dimensionality reduction cannot be estimated exactly, because it is related with smoothing the brightness of lungs, nevertheless it allows us to exclude contours of bones which confuse the search of nodules.

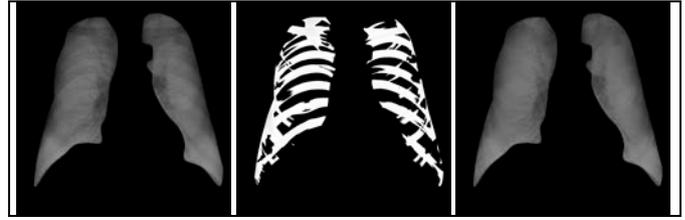

Fig. 7. Examples of the segmented JSRT image (left), its internal bone mask (central), and the result after bone shadows exclusion (right).

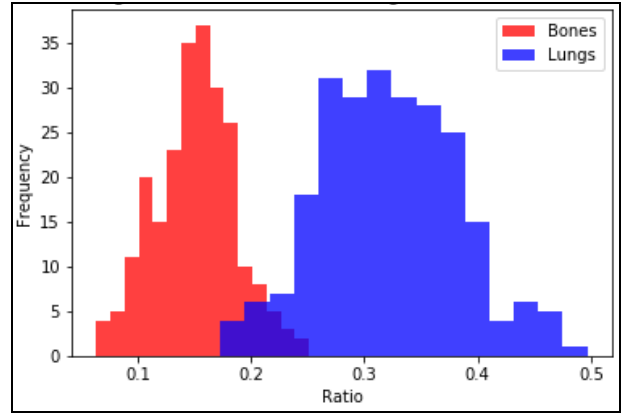

Fig. 8. Distribution of areas for lung masks (Fig.6) and bone masks (Fig.7).

### C. t-SNE for exclusion of outlying lung masks

The other application of dimensionality reduction was used for search and exclusion of the most dissimilar (outlying) lung masks (Fig.6, left) which occur due to variability of lungs and errors during manual lung segmentation. For this purpose the well-known t-distributed stochastic neighbor embedding (t-SNE) technique was applied [14,15]. It allowed us to embed high-dimensional lung mask image data into a 3D space, which can then be visualized in a scatter plot (Fig.9).

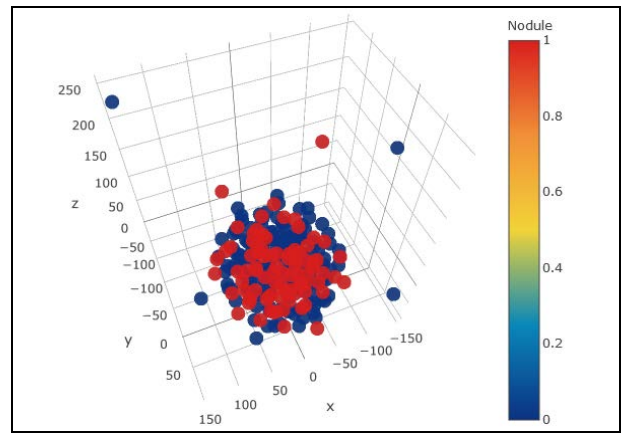

Fig. 9. Visualisation of t-SNE model for high-dimensional lung mask by a three-dimensional point, where similar lung mask are modeled by nearby points and dissimilar objects are mapped to distant points. Legend: red color – masks for images with nodules, blue – without nodules.

Actually, it maps each high-dimensional lung mask on a 3D (Fig.9) or 2D (Fig.10) point in a space where similar objects are modeled by nearby points and dissimilar objects are modeled by distant points.

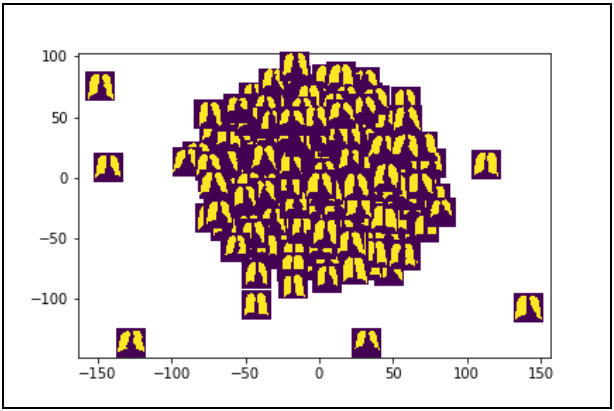

Fig. 10. Visualisation of lung mask images with the distance-preserving transformation of t-SNE from Fig.9 to get an impression of the most dissimilar (outlying) lung mask images.

By means of this dimensionality reduction on the basis of t-SNE technique the two types of the most dissimilar (outlying) lung masks were detected. The one outlier type is related with intrinsic variability of lungs (Fig.11, left), where lungs were abnormally small. The other outlier type is caused by errors during manual lung segmentation (Fig.11, right), where a heart region was not excluded from the left lobe (shown on the right white region, because the left lobe is shown on the right side).

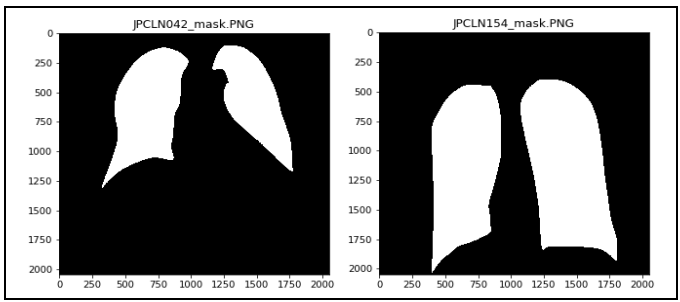

Fig. 11. Examples the most dissimilar (outlying) lung mask images obtained due to variability of lungs (left) and errors during manual lung segmentation (right), where heart region was not excluded from the left lobe (shown on the right white region).

Finally, by means of such search and exclusion of the most outlying lung masks (5% of all images), the additional dataset #5 was obtained and used for training and validating.

## IV. RESULTS

In this section, the results are presented as to effect of bone elimination, lung segmentation, and exclusion of outliers by t-SNE method on training and validation with regard to: the original JSRT dataset (below it is mentioned as dataset #01), BSE-JSRT dataset, i.e. the same JSRT dataset, but without clavicle and rib shadows (dataset #02), original JSRT dataset after segmentation (dataset #03), the BSE-JSRT dataset after segmentation (dataset #04), and the BSE-JSRT dataset after segmentation and exclusion of outliers by t-SNE method (dataset #05). The simple CNN [5] was trained in GPU mode (NVIDIA Tesla K40c card) by means of Tensorflow machine learning framework [16] on the 5 datasets to predict presence (154 images) or absence (93 images) of nodule.

The previous results on small versions of JSRT images (256*256) [5] demonstrated the drastic difference in training and validation results on the raw original images from JSRT database, and any of the pre-processed datasets. The CNN used in that previous work could not be trained on the original JSRT images (because of low image size and negligibly small nodule size), but it could be trained on the pre-processed (segmented and without shadows of bones) images. Here in continuation of this former work the same CNN was applied to much bigger (2048*2048) images to check the influence of the similar preprocessing and filtering out outliers (i.e. CXR images with outlying masks) by dimensionality reduction on the basis of t-SNE technique.

Several training and validation runs for the CNN on CXR images from JSRT database (dataset #1) were performed. The examples of some of these runs are shown in Fig.12 (color symbols) along with the results of their averaging over the runs (Fig.12, red line) and smoothing the averaged time series (Fig.12, blue line) by locally weighted polynomial regression method [17].

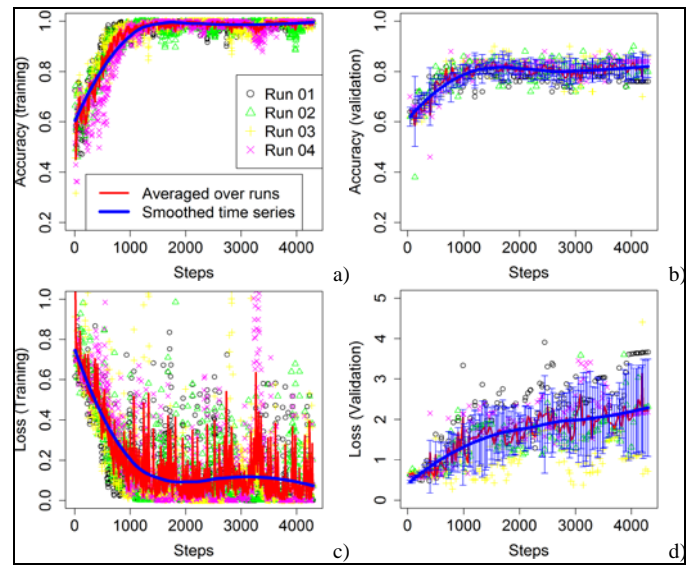

Fig. 12. Example of some cross-validation runs for the CXR images from JSRT database (dataset #1): training accuracy (a), validation accuracy (b), training loss (c), and validation loss (d) (different colors correspond to different runs).

The evident over-training can be observed after comparison of training and validation results, where the averaged and smoothed value of training accuracy is going with epochs to the theoretical maximum of 1 and training loss is going to 0, while the averaged and smoothed value of validation accuracy is no more than 0.8 with the very high and increasing values of validation loss.

The similar several training and validation runs for the CNN were performed on other datasets mentioned above. Below the examples of some of these runs on the CXR images without bone shadows from BSE-JSRT dataset after lung segmentation and filtering out the CXR images with the outlying lung masks (dataset #5) are shown in Fig.13 (color symbols) along with the results of their averaging over the runs (Fig.13, red line) and smoothing the averaged time series (Fig.13, blue line) by locally weighted polynomial regression method [17].

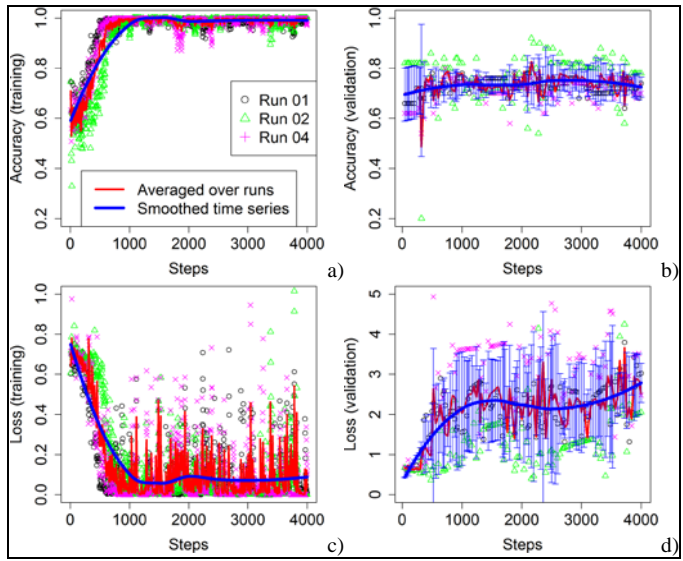

Fig. 13. Example of some cross-validation runs on the CXR images without bone shadows from BSE-JSRT dataset after lung segmentation and filtering out the CXR images with the outlying lung masks (dataset #5): training accuracy (a), validation accuracy (b), training loss (c), and validation loss (d) (different colors correspond to different runs).

For all of these datasets the similar over-training was observed after comparison of the averaged and smoothed training and validation values of accuracy (Fig.14a) and loss (Fig.14c).

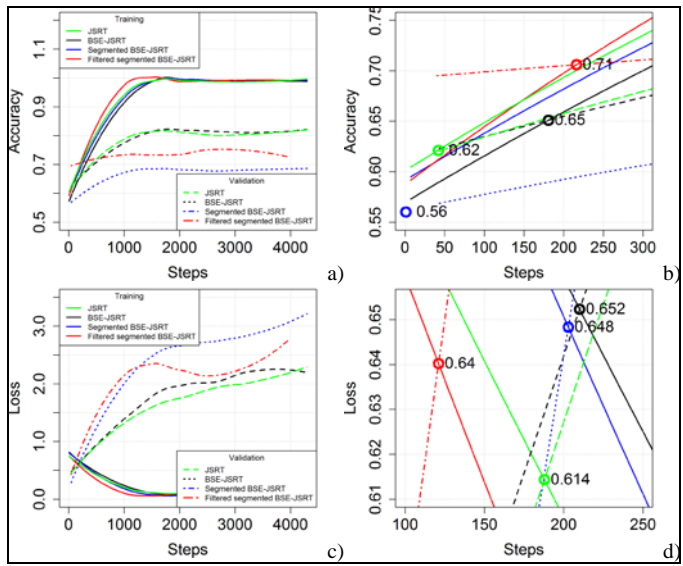

Fig. 14. Averaged and smoothed values of accuracy (a) with an increased view on them (b), and averaged and smoothed values of loss (c) with an increased view on them (d) for all datasets noted in the legend.

The closer look allow us to find the stages where training values of accuracy (Fig.14b) and loss (Fig.14d) become equal or higher than their validation counterparts that allow us to estimate the actual values of accuracy and loss.

V. DISCUSSION AND CONCLUSIONS

The careful analysis of training and validation values of accuracy and loss (Fig.14) for images of the highest possible sizes (2048*2048) from the original dataset #01 and pre-processed datasets #02, #04, #05 allows us to derive the following observations.

The training rate is highest for the most thoroughly pre-processed dataset #5 obtained after lung segmentation, bone shadow exclusion, and filtering out the outliers. It is lowest for the pre-processed dataset #4 obtained after lung segmentation and bone shadow exclusion, but without filtering out the outliers. This difference could be explained by aforementioned exclusion (dataset #5) and availability (dataset #4) of outliers due to intrinsic variability of lungs (Fig.11, left) and errors during manual lung segmentation (Fig.11, right). The similar tendency was observed for the estimated actual value of accuracy (Fig.14b), which is highest (0.71) for the most thoroughly pre-processed dataset #5 and lowest for dataset #4 (<0.56).

Despite the high dimensionality reduction the bone shadow exclusion technique applied to get the dataset #02 does not give the significant improvement of accuracy (0.65) in comparison (Fig.14b) to the accuracy obtained on the original dataset #01 (0.62). It is probably caused by availability and influence of the outliers mentioned before, which can be effectively excluded by t-SNE dimensionality reduction method with the significant increase of accuracy even for the very low portion of the filtered our outliers (5%) and the very simple CNN used here and in our previous similar work [5].

The results obtained here demonstrate that the pre-processing techniques like bone shadow elimination and segmentation can improve the accuracy, if they will be used along with exclusion of outliers by t-SNE method in such simplified configuration even (simple CNN and small not well-balanced dataset). The reverse effect (decrease of accuracy) can be observed if some complicated pre-processing technique will be used which lead to appearance of additional artifacts and correspondent outliers (like very different shapes of the lungs and lung border patterns).

It should be noted that further improvement in detection by CXR of marks of these lung diseases could be obtained by:

- exclusion of outliers by t-SNE or other dimensionality reduction method after every pre-processing stage where additional artifacts can appear;
- increase of the investigated datasets from the current small number of 247 images (JSRT and BSE-JSRT datasets) up to >1000 (MC dataset) and > 100 000 images (ChestXRay dataset), and the significant progress can be reached by open sharing the numerous similar datasets from hospitals around the world in the spirit of the open science data, volunteer data collection, data processing and computing [18];

- data augmentation with regard to lossy and lossless transformations which under work now [19].

In conclusion, the results obtained here demonstrate the efficiency of the considered pre-processing techniques of dimensionality reduction (especially by t-SNE for filtering out the outliers) for the simple CNN and small not well-balanced dataset even. It should be emphasized that the pre-processed dataset obtained after lung segmentation, bone shadow exclusion, and filtering out the outliers by t-SNE (dataset #05) demonstrates the highest training rate and best accuracy in comparison to the other pre-processed datasets after bone shadow exclusion (dataset #02) and lung segmentation (dataset #04). That is why the additional reserve of development could be related with:

- the further dimensionality reduction by other limb (like heart, arms, etc.) shadow elimination on the basis of the more complicated semantic segmentation techniques,
- increase of size and complexity of the deep learning CNN from the current size of <10 layers (up to >100 layers like in the current most accurate networks like CheXNet [4]) and fine tuning the deep learning models will be very important, because it can be crucial for efficiency of the models used [20-22].

## Acknowledgment

The work was partially supported by Huizhou Science and Technology Bureau and Huizhou University (Huizhou, P.R.China) in the framework of Platform Construction for China-Ukraine Hi-Tech Park Project # 2014C050012001.